\documentclass[11pt]{article}

\usepackage[preprint]{acl}

\usepackage{times}
\usepackage{latexsym}

\usepackage[T1]{fontenc}

\usepackage[utf8]{inputenc}
\usepackage{amsmath}
\usepackage{amssymb}
\usepackage{enumitem}
\usepackage{booktabs}
\usepackage{multirow}   
\usepackage{array}     
\usepackage{algorithm}
\usepackage{algpseudocode}
\usepackage{microtype}

\usepackage{inconsolata}
\usepackage{url}
\usepackage{graphicx}
\usepackage{xcolor}

\title{Beyond Fresh Starts: Stateful Inference for Streaming ASR in Conversational Voice Agents}

\author{Sameep Chattopadhyay$^\dagger$ \, Alexander Erdmann$^\ddagger$ \, Mari Ostendorf$^\dagger$ \\
  $^\dagger$\,University of Washington \, $^\ddagger$\,SRI International \\
  \texttt{sameepch@uw.edu \, alex.erdmann@sri.com \, ostendor@uw.edu} \\}

\begin{document}
\maketitle
\begin{abstract}

Modern voice-agent systems rely on streaming speech recognition models that operate under stringent latency constraints.  This study shows that, due to the limited memory constraints of real-time processing, these systems are adversely impacted by conversational phenomena such as long silences and backchannels. While many agentic pipelines mitigate this by resetting state at each turn, this approach discards vital context and impairs performance at turn onsets.
We propose two state-management strategies that preserve cross-utterance context to reduce onset errors.
In experiments with two state-of-the-art streaming models on two spoken dialogue benchmarks, 
our best method yields an average of 15--21\% relative WER reduction at utterance onsets.


\end{abstract}

\section{Introduction}

The past few years have seen a rapid expansion in conversational voice agents. In 2026, the vast majority of deployed voice agents follow a cascaded speech-to-text, language model, and text-to-speech 
pipeline \cite{hamming2025latency}.
To mimic human turn-taking dynamics, these systems have to operate under tight latency 
budgets, with response delays below 500\,ms \cite{stivers2009universals}. Industry benchmarks suggest that ASR alone should contribute no more than 200\,ms to the overall pipeline \cite{hamming2025latency}, thus requiring streaming ASR, i.e., models that transcribe concurrently as the speaker talks.

Unlike offline systems that can utilize bidirectional global contexts, streaming ASR models must operate with very limited acoustic memory, often in the range of a few seconds \cite{rekesh2023fastconformer,defossez2024moshi} to support real-time processing. The long silences and frequent backchannels (\emph{e.g.}, ``mm-hm'', ``uh-huh'') in conversational speech pose a challenge to such low-latency models, by displacing crucial acoustic information from the prior utterance with more recent but task-irrelevant content \cite{hsiao2020online}.
\begin{figure}[t]
    \centering
    \includegraphics[width=\columnwidth]{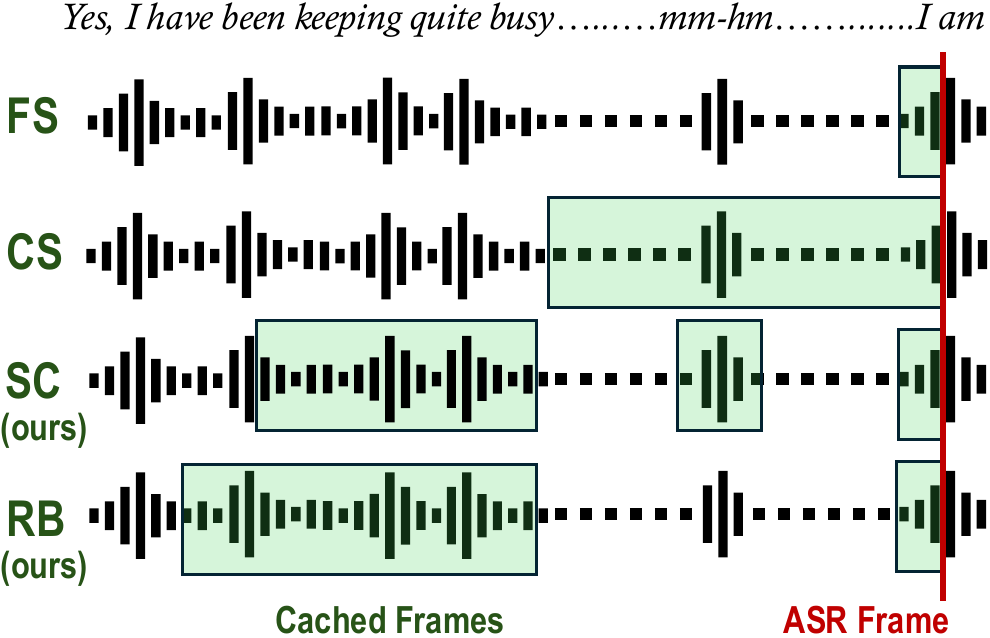}
    \vspace*{-1em}
    \caption{\textbf{Stateful inference.} Cached acoustic memory state for streaming ASR models under the  different state-management strategies defined in Section \ref{subsec:state}.}
    \label{fig:state_plot}
    \vspace*{-1.5em}
\end{figure}
Most agentic pipelines handle this by simply resetting the state at each utterance turn, giving the model a fresh start. Useful prior context is then lost, sometimes causing the ASR model to falter on tokens at the beginning of a turn. {Such errors may also affect downstream language-model components, as prior work suggests that LLMs often place disproportionate weight on early input tokens during inference \cite{Xiao2024EfficientSL}.} 

In this work, we focus on the FastConformer RNN-Transducer (RNNT) 
architecture \cite{noroozi2024stateful}, a {widely adopted architecture} for low-latency streaming ASR. We first show that these models are 
susceptible to performance degradation caused by long pauses, 
backchannels, and state resets at turn boundaries. We then propose 
a set of simple state-management strategies\footnote{Code: \url{https://github.com/Sameep-c/stateful-streaming-asr.git}} to mitigate these effects 
while preserving prior utterance context across turns. Evaluating on two of the best-performing low-latency streaming ASR models across two telephonic 
dialogue datasets, we achieve a 15--21\% relative WER 
reduction at turn onsets over the standard state-resetting baselines.

\section{Streaming ASR}
\label{sec:bg}
Deployed streaming ASR systems broadly fall into two inference paradigms. Live 
captioning and real-time transcription services generally process audio 
as a  continuous stream from start to finish, relying on the model's 
memory to carry context across the entire recording 
\cite{machavcek2023whisper, raj2022continuous}. In contrast, conversational voice agents used in dialogue settings pair streaming ASR with voice activity detection (VAD) or turn detection and decode each turn independently. In other words, the model 
state is reset at every utterance boundary \cite{huang2022e2e, 
narayanan2020recognizing}.

The two frameworks have distinct failure modes for conversational speech. 
Continuous streaming accumulates information across the entire recording, including extended silences and non-lexical backchannels, thereby corrupting acoustic representations
of subsequent speech \cite{hsiao2020online,choi2024joint}. Dialogue agents avoid this problem by restarting the model at each turn, but at the cost of lost acoustic and linguistic 
context established in earlier turns. Previous work has shown that this particularly degrades performance in long-form RNN-T decoding \cite{narayanan2020recognizing, cast2023, hou2022dialogue}.

Approaches for utilizing prior context have primarily focused on 
injecting text-based dialogue history into the prediction network 
\cite{hou2022dialogue} or using audio context from prior segments to train 
the encoder \cite{masumura2022contextual,schwarz2021rnnt}. The closest work to ours is \citet{narayanan2020recognizing}, who show that 
carrying LSTM encoder states across utterances during training helps models 
generalize to long-form speech, serving as a workaround for the practical 
constraint of training on short utterances. In contrast, 
our methods do not require any training or fine-tuning. Instead, we approach
state management as an inference-time decision problem, using streaming models already trained on long-form audio. 

Among open-source streaming ASR models, the cache-aware FastConformer-RNNT 
\cite{noroozi2024stateful} has become the go-to architecture for sub-500\,ms 
latency applications. Streaming is enabled via cache-aware attention, where each encoder frame attends to a fixed left-context window of $L$ frames. These key-value (KV) pairs are stored in a rolling cache alongside a configurable right-context of $R$ lookahead frames. In addition, each encoder layer maintains a localized convolution cache to mitigate potential artifacts at chunk boundaries.
The decoder follows the RNN-Transducer formulation \citep{graves2012sequence}, wherein
the prediction network is a single-layer LSTM operating over the sequence of previously emitted
non-blank tokens with its hidden and cell states 
acting as the decoder's linguistic memory \cite{noroozi2024stateful}.

\section{Study Design}
\label{sec:study}
{This section presents the experimental setup, including the streaming ASR models, datasets, and evaluation protocol. It also formalizes the proposed state-management strategies, State Carry-over and State Rollback, along with the baseline methods used for comparison.}
\subsection{Models}
We evaluate the following pre-trained streaming ASR models: \texttt{fastconformer-114m} \cite{noroozi2024stateful} and \texttt{nemotron-streaming-0.6b} \cite{nemotron2024streaming}. 
Both models share a common cache-aware FastConformer-RNNT architecture, with model state given by 
a set of six  components distributed between the encoder and the decoder. The FastConformer state comprises
the encoder attention KV cache,
a per-layer convolution cache, 
and the corresponding valid cache length. Similarly, the RNNT state consists of
the accumulated decoder hypothesis,
the last predicted token, and
the number of tokens emitted.
Further architectural details are provided in Appendix \ref{app:arch}.
\subsection{Datasets} We evaluate our framework on the test sets of two standard telephonic dialogue benchmarks: CallHome \cite{canavan1997callhome} and Switchboard \cite{godfrey1992swb,Deshmukh1998ResegmentationOS}
(test split as in \citet{romana2023}). For both datasets, we use individual (unmixed) channels, which contain backchannels, extended periods of silence, and occasional crosstalk. 
The datasets provide utterance-level timestamps for each speaker. 
Consecutive utterances from the same speaker separated by less than 1\,s of silence are treated as a single turn, consistent with end-of-turn detection heuristics employed by commercial voice agents \cite{hamming2025latency}. Both models are pre-trained on the Switchboard training set and have not been 
exposed to CallHome during training \cite{noroozi2024stateful,nemotron2024streaming}. Additional dataset details are provided in Appendix \ref{app:data}.


\subsection{State Management}
\label{subsec:state}
 We implement and compare the four state management strategies visualized in Figure \ref{fig:state_plot}:

\begin{itemize}[noitemsep, topsep=0pt, leftmargin=*]

    \item \textit{Fresh Start (FS):} The model's internal state is fully 
    reset at the start of each turn, causing each utterance to be decoded 
    independently.
    
    \item \textit{Continuous State (CS):} Streaming proceeds continuously without any manipulation. The acoustic memory always consists of the last $L$ frames from the speaker channel, regardless of content.

    \item \textit{State Carry-over (SC):}  The state at the end of a turn is utilized as the starting state for the next one, allowing the model to retain context from previous turns while bypassing long silences.

    \item \textit{State Rollback (RB):} An extension of state carry-over in which the model rolls back
    to the state saved at the end of the last substantial turn when the previous turn is too short ( $< k$\textit{ tokens}) in order to drop backchannel context.
\end{itemize}

{Algorithm~\ref{alg:state_mgmt} formalizes the three state management strategies (FS, SC, RB) as a single state-update 
rule over speaker turns. CS, which streams continuously through inter-turn silence without any manipulation, 
is excluded from the algorithm.}
\vspace*{-0.5em}
\begin{algorithm}[h]
\caption{State Management Strategies for Streaming ASR}
\label{alg:state_mgmt}
\footnotesize
\begin{algorithmic}[1]
\Require Audio stream with turns $\{u_1, \dots, u_T\}$; mode $m \in \{FS, SC, RB\}$; threshold $k$ (for RB)
\vspace*{0.2em}
\Ensure Hypotheses $\{h_1, \dots, h_T\}$, where $h_i$ is the decoded token sequence for turn $u_i$ and $|h_i|$ is its token length 
\vspace*{0.2em}
\State $S \gets \varnothing$ \Comment{Model state}
\State $S^{*} \gets \varnothing$ \Comment{Last substantial-turn state (RB)}
\For{$i = 1$ to $T$}
   \State $h_i, S_i\gets \textsc{Infer}(u_i, S)$
    \If{$m = FS$}
        \State $ S \gets  \varnothing$  \Comment{State reset}
    \ElsIf{$m = SC$}
        \State $ S \gets  S_{i}$  \Comment{State carry-over}
    \ElsIf{$m = RB$}
            \If{ $|h_i| \geq k$} 
            \State $S \gets  S_{i}$
            \State $S^* \gets S_i$  
            \Else
            \State $ S \gets  S^*$  \Comment{State rollback}
        \EndIf
    \EndIf
\EndFor
\State \Return $\{h_1, \dots, h_T\}$
\end{algorithmic}
\end{algorithm}
\vspace*{-0.5em}

Most commercially deployed voice agents operate in a manner similar to FS, while for other streaming ASR applications, like live meeting captioning, streaming happens without any state manipulation like CS. As described in Section \ref{sec:bg}, both  FS and CS
have shortcomings, which the SC and RB strategies are designed to address.

\begin{figure*}[t]  
    \centering
    \includegraphics[width=\textwidth]{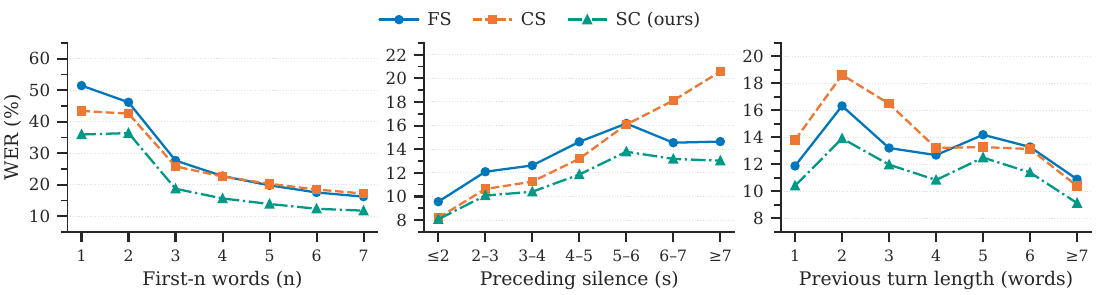} 
    \vspace*{-1.5em}
    \caption{Word Error Rates for different state-management strategies with \texttt{fastconformer-114m} on Switchboard.}
    \label{fig:results_plot}
\vspace*{-0.5em}
\end{figure*}

\begin{table*}[t]
\centering
\small
\begin{tabular}{ll cccc cccc}
\toprule
\multirow{2}{*}{\textbf{Model}} & \multirow{2}{*}{\textbf{Category}} & \multicolumn{4}{c}{\textbf{CallHome}} & \multicolumn{4}{c}{\textbf{Switchboard}} \\
\cmidrule(lr){3-6} \cmidrule(lr){7-10}
& & \textsc{fs} & \textsc{cs} & \textsc{sc} & \textsc{rb} & \textsc{fs} & \textsc{cs} & \textsc{sc} & \textsc{rb} \\
\midrule
\multirow{4}{*}{\texttt{fastconformer-114m}} 
& Overall & 15.43 & 15.18 & 14.02 & \textbf{13.83} & 11.97 & 12.36 & 10.26 & \textbf{10.20} \\
& First Word & 48.10 & 41.40 & 36.24 & \textbf{35.03} & 51.44 & 44.32 & 35.92 & \textbf{34.78} \\
& After Long Pause & 17.17 & 20.66 & 16.15 & \textbf{15.70} & 14.73 & 20.06 & 13.08 & \textbf{12.99} \\
& After Short Turn & 18.52 & 20.20 & 17.17 & \textbf{16.61} & 13.93 & 15.94 & 12.03 & \textbf{11.94} \\
\midrule
\multirow{4}{*}{\texttt{nemotron-streaming-0.6b}} 
& Overall & 13.82 & 14.29 & 13.86 & \textbf{13.52} & 10.90 & 11.40 & 10.85 & \textbf{10.62} \\
& First Word & 32.28 & 32.61 & 32.36 & \textbf{27.85} & 32.50 & 34.40 & 31.69 & \textbf{29.46} \\
& After Long Pause & 14.79 & 16.64 & 15.62 & \textbf{14.68} & 13.25 & 14.37 & 13.11 & \textbf{12.83} \\
& After Short Turn & 16.64 & 18.31 & 17.05 & \textbf{15.66} & 12.51 & 13.62 & 12.62 & \textbf{12.15} \\
\bottomrule
\end{tabular}
\caption{Detailed WER (\%) analysis. \textit{First Word} refers to the WER of the initial word in a turn; ``Long Pause'' denotes silences exceeding the acoustic memory length of 5.6\,s; ``Short Turn'' denotes turns shorter than 3 words.}
\label{tab:phenomena_analysis_v3}
\vspace*{-1.5em}
\end{table*}

\subsection{Evaluation}
We evaluate our state-management strategies using the standard Word Error Rate (WER) measure. For all experiments, the models perform greedy decoding with a batch size of 1, operating with a left-context window of 
$L = 70$ frames (5.6\,s of history) and a right-context of $R = 1$ frame 
(160\,ms latency), following the configuration described in \citet{noroozi2024stateful}. 
This places the models within the latency requirements of real-time 
voice agents. For RB, we experiment with token thresholds  $k \in \{1, 5, 10\}$, and {present results using the hyperparameters 
that achieve the best validation-set performance: $k=5$ for \texttt{fastconformer-114m} and $k=10$ for 
\texttt{nemotron-streaming-0.6b}.}

For the key experiments reported in Section \ref{sec:res}, we use the ground-truth timestamps provided by each dataset to identify spoken regions and filter out tokens generated outside these intervals, such as those arising from crosstalk. To accommodate late token emissions that may occur past the utterance boundary, we extend each spoken region by 500\,ms beyond the annotated end time for all our experiments. {We further evaluate our strategies in a more realistic deployment setting using turn boundaries generated by Silero-VAD \cite{silero2021vad}, a voice activity detection model widely used in commercial voice agents \cite{hamming2025latency}. Implementation details including VAD hyperparameters are provided in Appendix \ref{app:vad}.}

{\subsection{Memory Overhead}
As described in Algorithm~\ref{alg:state_mgmt}, SC incurs no additional memory 
overhead beyond standard streaming inference, as it simply carries forward the 
model state produced at the end of the previous turn. In contrast, RB needs to store
an additional state $(S^*)$ from the last substantial turn. For the default configuration described above, this amounts to approximately
2.6\,MB for \texttt{fastconformer-114m} and 7.32\,MB for
\texttt{nemotron-streaming-0.6b} (see Appendix~\ref{app:arch}). This corresponds
to just 0.6\% and 0.3\% of the respective model sizes, making the additional
memory overhead of RB negligible.}

\section{Results and Analysis}
\label{sec:res}

We first conduct a set of experiments to characterize the failure modes of different state-management strategies under varying conversational conditions. Specifically, we measure WER as a function of (i) the first $n$ words in the turn (for turns with $\ge n$ words); (ii) the duration of silence preceding the turn; and (iii) the length of the preceding turn. The first analysis  evaluates performance at turn onsets, while the latter two assess the robustness of the strategies to conversational noise.

Figure \ref{fig:results_plot} illustrates the performance of the FS, CS, and SC strategies for these failure modes using \texttt{fastconformer-114m} on Switchboard.
The first plot highlights that resetting the state results in significant performance degradation in the absence of any prior context for the first few words.  
The second plot shows how the preceding silence can degrade the state. While CS is generally better than FS for turns following shorter silences, CS performance degrades when the silence length exceeds the acoustic memory length of 5.6\,s. 
The third plot suggests that short preceding turns, which often correspond to backchannels, degrade CS performance on the subsequent utterance relative to FS. 
In all contexts, the proposed SC approach outperforms existing baselines, demonstrating how state management through SC can effectively preserve relevant cross-utterance context.

\begin{figure}[h]
    \centering
    \includegraphics[width=\columnwidth]{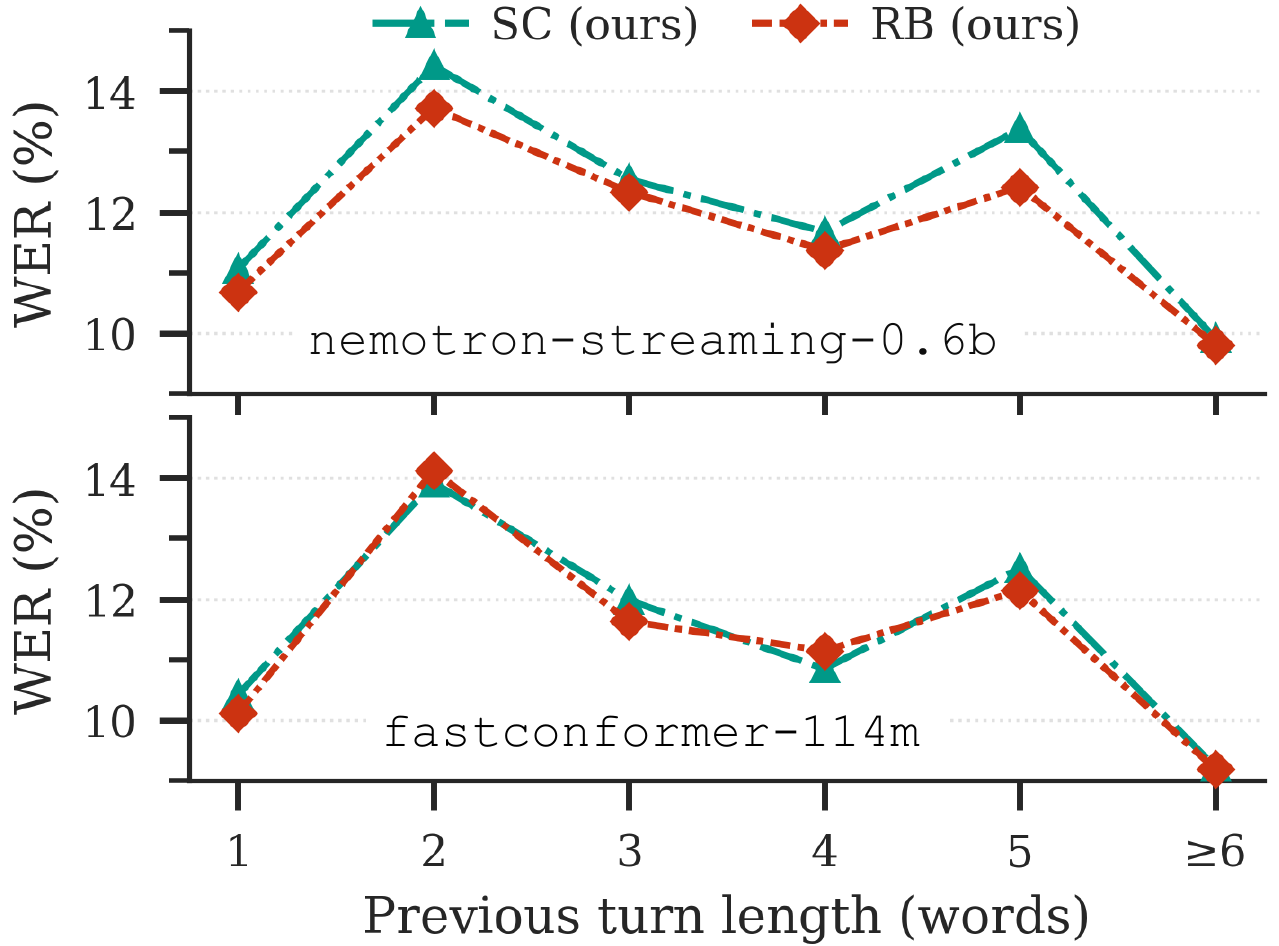}
    \caption{Comparison of SC and RB for Switchboard.}
    \label{fig:RB_plot}
    \vspace*{-1em}
\end{figure}

 Figure~\ref{fig:RB_plot} shows the WER of SC compared to RB for utterances after turns of different lengths, for both models on Switchboard. 
For \texttt{fastconformer-114m}, RB and SC perform similarly. However, RB provides noticeably larger improvements for \texttt{nemotron-streaming-0.6b}, suggesting that its model state is more sensitive to shorter turns and backchannels.

All four strategies are compared in Table \ref{tab:phenomena_analysis_v3} for all model and dataset pairs. The key findings from the plots in Figure~\ref{fig:results_plot} are captured in the following WER metrics:
 i)~\textit{Overall:} the primary WER.
ii)~ \textit{First Word:} WER comparing the first word in the reference turn and the ASR hypothesis.
iii)~\textit{After Long Pause:}  WER of turns preceded by pauses  $\geq5.6$ s.
iv)~ \textit{After Short Turn:} WER of utterances following short turns with $\leq 2$ words (note that the vast majority of these short turns are backchannels). 

Table \ref{tab:phenomena_analysis_v3} shows that RB achieves the lowest overall WER across all models and datasets, demonstrating the benefits of selectively preserving cross-utterance context. In contrast, naive CS performs worst in three of four settings, largely due to errors following long pauses and backchannels.
While FS mitigates these effects by resetting state at each turn, it incurs a substantial turn-onset penalty, particularly for \texttt{fastconformer-114m}, where first-word WER degrades despite improvements in overall WER. Preserving prior context via SC provides clear gains for \texttt{fastconformer-114m}, but 
not for \texttt{nemotron-streaming-0.6b}, where utterances following backchannels often suffer degradation. RB addresses this issue by selectively discarding harmful context, consistently outperforming both FS and SC across all evaluated metrics.

{\paragraph{Statistical significance.} To verify that the improvements from our methods are statistically significant, we compute ``Overall WERs'' for FS and RB and compare them using the MAPSSWE test, following standard practice in the ASR literature \cite{gillick1989some}. Our analysis reveals that, across all four model-dataset combinations, RB's improvement over FS is statistically significant, with  $p<10^{-3}$ for three setups and $p<0.05$ for the remaining one
(Appendix \ref{app:sigtest}). While the overall gains appear modest, this is expected: the benefits from our strategies are concentrated in the 
first few tokens of each turn and are averaged out over the complete utterances. The first-word WER 
results in Table~\ref{tab:phenomena_analysis_v3} directly capture this effect.}

{ \paragraph{Rollback threshold.} We experiment with coarse thresholds $k \in \{1, 5, 10\}$ for RB, and present results with the best hyperparameters, i.e., $k=5$ for 
\texttt{fastconformer-114m} and $k=10$ for \texttt{nemotron-streaming-0.6b} in Table \ref{tab:phenomena_analysis_v3}. Additional results with the other 
thresholds are provided in Appendix~\ref{app:eval}. Based on these results, we observe that RB consistently 
outperforms SC across all tested values of $k$.}

\paragraph{Variable latency.} The performance trends of the state-management strategies remain largely unchanged even with different latencies, with RB consistently achieving the best results at all supported latencies below 250 ms (Appendix \ref{app:latency}). 
 Reducing latency from 560\,ms to 160\,ms to satisfy voice-agent requirements results in only a modest degradation in performance. For \texttt{fastconformer-114m}, RB at 160\,ms achieves a lower WER than FS at 560\,ms. With RB, across both models, the WER increase relative to the best-performing strategy at 560\,ms remains below 6\%.

\paragraph{Without ground-truth timings.} Finally, we evaluate our methods using turn boundaries generated by Silero-VAD \cite{silero2021vad}, to better approximate real-world deployment conditions. The results (Appendix \ref{app:vad}) show that our strategies continue to outperform existing approaches without access to ground-truth timings, with the improvements being statistically significant.  At 160\,ms latency, RB achieves a 20.6\% relative reduction in first-word WER over FS using VAD-obtained boundaries for Switchboard, closely matching the 20.9\% reduction observed in Table \ref{tab:phenomena_analysis_v3}. This analysis is restricted to the Switchboard dataset, as it requires word-level timing annotations for turn attribution, which are not available in CallHome.

To summarize, our analysis yields the following key findings:
\begin{itemize}[noitemsep, topsep=0pt, leftmargin=*]
\item Although FS outperforms CS in most contexts, 
it consistently incurs a substantial penalty on the initial words at each turn.
\item CS 
performance degrades as the preceding silence duration increases.
\item SC substantially outperforms CS on both models and benchmarks. The gains from SC over FS are mainly for \texttt{fastconformer-114m}.
\item RB is the best-performing state-management strategy across all evaluated models and datasets.
\item These trends are consistent across all latencies below 250\,ms  and hold even with VAD-obtained turn boundaries for Switchboard.
\end{itemize}

\section{Conclusions}
This work provides a detailed empirical analysis of how long silences and backchannels degrade low-latency 
streaming ASR. We demonstrate that the state-resetting approach 
used by most voice agents discards useful cross-utterance context, 
causing disproportionate performance degradation at utterance onsets 
where errors are most costly for the downstream language model. 
To address this, we propose two simple state-management strategies, 
State Carry-over (SC) and State Rollback (RB), that preserve prior 
context across turns while remaining robust to 
conversational noise. Experiments across two streaming 
ASR models and two conversational benchmarks confirm that our strategies  improve overall recognition accuracy, especially at turn onsets. 
These improvements can be particularly important for subsequent downstream processing by voice agents, since early ASR errors can propagate to and compound in downstream components.
\newpage
\section{Limitations}
Our experiments are limited to benchmarks consisting primarily of American English telephonic conversations; whether the observed trends generalize to other languages and conversational settings remains an open question.  The proposed rollback strategy relies on an empirically tuned threshold. While we demonstrate consistent gains across varied settings, adapting this threshold dynamically to conversational context may yield further improvements. Furthermore, this work focuses on a specific set of conversational failure modes: state resets, long pauses, and backchannels. It would be useful to extend the analysis to additional challenges, such as noisy acoustic conditions, code-switching, and multi-speaker settings. {Finally, this work focuses on improving ASR performance rather than evaluating downstream voice-agent behavior. Although improved recognition may benefit subsequent language-model components, we do not directly measure its impact on downstream tasks such as intent prediction, dialogue quality, or task success. Investigating ASR error propagation in LLM-based voice-agent pipelines remains an important direction for future work.} 
\section*{Acknowledgments}

This work was supported by the Intelligence Advanced Research Projects Activity (IARPA) via Department of Interior/Interior Business Center (DOI/IBC) contract number 140D0424C0070. The United States
Government is authorized to reproduce and distribute reprints for governmental purposes, notwithstanding any copyright annotation thereon. Disclaimer: The views and conclusions contained herein are those of the authors and should not be interpreted as necessarily representing the official policies or endorsements, either expressed or implied, of IARPA, DOI/IBC, or the U.S. Government.
\bibliography{custom}

@String{Computer = "{IEEE} Computer" }

@inproceedings{rekesh2023fastconformer,
  title     = {Fast {Conformer} with Linearly Scalable Attention for
               Efficient Speech Recognition},
  author    = {Rekesh, Dima and Koluguri, Nithin Rao and Kriman, Samuel
               and Majumdar, Somshubra and Noroozi, Vahid and Huang, He
               and Hrinchuk, Oleksii and Puvvada, Krishna and Kumar, Ankur
               and Balam, Jagadeesh and Ginsburg, Boris},
  booktitle = {2023 {IEEE} Automatic Speech Recognition and Understanding
               Workshop ({ASRU})},
  pages     = {1--8},
  year      = {2023},
  doi       = {10.1109/ASRU57964.2023.10389701},
  eprint    = {2305.05084},
  archivePrefix = {arXiv},
}

@inproceedings{noroozi2024stateful,
  title={Stateful Conformer with Cache-Based Inference for Streaming Automatic Speech Recognition},
  author={Vahid Noroozi and Somshubra Majumdar and Ankur Kumar and Jagadeesh Balam and Boris Ginsburg},
  journal={ICASSP 2024 - 2024 IEEE International Conference on Acoustics, Speech and Signal Processing (ICASSP)},
  year={2023},
  pages={12041-12045},
  doi={10.1109/ICASSP48485.2024.10446861}
}

@article{graves2012sequence,
  title   = {Sequence Transduction with Recurrent Neural Networks},
  author  = {Graves, Alex},
  journal = {arXiv preprint arXiv:1211.3711},
  year    = {2012},
  eprint  = {1211.3711},
  archivePrefix = {arXiv},
  primaryClass  = {cs.NE},
  doi     = {10.48550/arXiv.1211.3711},
}

@techreport{defossez2024moshi,
  title         = {{Moshi}: a speech-text foundation model for real-time
                   dialogue},
  author        = {D{\'e}fossez, Alexandre and Mazar{\'e}, Laurent
                   and Orsini, Manu and Royer, Am{\'e}lie
                   and P{\'e}rez, Patrick and J{\'e}gou, Herv{\'e}
                   and Grave, Edouard and Zeghidour, Neil},
  year          = {2024},
  eprint        = {2410.00037},
  archivePrefix = {arXiv},
  primaryClass  = {eess.AS},
  url           = {https://arxiv.org/abs/2410.00037},
}

@inproceedings{schwarz2021rnnt,
  title     = {Improving RNN-T ASR Accuracy Using Context Audio},
  author    = {Schwarz, Andreas and Sklyar, Ilya and Wiesler, Simon},
  booktitle = {Proc. Interspeech 2021},
  pages      = {1792--1796},
  year       = {2021},
  doi        = {10.21437/Interspeech.2021-542}
}

@article{hsiao2020online,
  author    = {Hsiao, Roger and Can, Dogan and Ng, Tim and Travadi, Ruchir and Ghoshal, Arnab},
  title     = {Online Automatic Speech Recognition With Listen, Attend and Spell Model},
  journal   = {IEEE Signal Processing Letters},
  volume    = {27},
  pages     = {1889--1893},
  year      = {2020},
  publisher = {IEEE},
  doi       = {10.1109/LSP.2020.3031480}
}

@article{Deshmukh1998ResegmentationOS,
  title={Resegmentation of SWITCHBOARD},
  author={Neeraj Deshmukh and Aravind Ganapathiraju and Andi Gleeson and Jonathan Hamaker and Joseph W. Picone},
  journal={5th International Conference on Spoken Language Processing (ICSLP 1998)},
  year={1998},
  url={https://api.semanticscholar.org/CorpusID:17232323}
}

@inproceedings{godfrey1992swb,
author = {Godfrey, John J. and Holliman, Edward C. and McDaniel, Jane},
title = {SWITCHBOARD: telephone speech corpus for research and development},
year = {1992},
isbn = {0780305329},
publisher = {IEEE Computer Society},
address = {USA},
booktitle = {Proceedings of the 1992 IEEE International Conference on Acoustics, Speech and Signal Processing - Volume 1},
pages = {517–520},
numpages = {4},
location = {San Francisco, California},
series = {ICASSP'92}
}

@misc{canavan1997callhome,
  author       = {Canavan, Alexandra and Graff, David and Zipperlen, George},
  title        = {{CALLHOME} American English Speech},
  howpublished = {Linguistic Data Consortium, Philadelphia},
  year         = {1997},
  note         = {LDC97S42},
  url={https://catalog.ldc.upenn.edu/LDC97S42}
}

@article{choi2024joint,
  title     = {Joint streaming model for backchannel prediction and automatic 
               speech recognition},
  author    = {Choi, Yong-Seok and Bang, Jeong-Uk and Kim, Seung Hi},
  journal   = {ETRI Journal},
  volume    = {46},
  number    = {1},
  pages     = {118--126},
  year      = {2024},
  publisher = {Wiley},
  doi       = {10.4218/etrij.2023-0358}
}

@inproceedings{hou2022dialogue,
  title     = {Bring dialogue-context into {RNN-T} for streaming {ASR}},
  author    = {Hou, Junfeng and Chen, Jinkun and Li, Wanyu and Tang, Yufeng 
               and Zhang, Jun and Ma, Zejun},
  booktitle = {Proc. Interspeech},
  pages     = {2048--2052},
  year      = {2022},
  doi       = {10.21437/Interspeech.2022-697}
}

@inproceedings{raj2022continuous,
  author={Raj, Desh and Lu, Liang and Chen, Zhuo and Gaur, Yashesh and Li, Jinyu},
  booktitle={ICASSP 2022 - 2022 IEEE International Conference on Acoustics, Speech and Signal Processing (ICASSP)}, 
  title={Continuous Streaming Multi-Talker ASR with Dual-Path Transducers}, 
  year={2022},
  volume={},
  number={},
  pages={7317-7321},
  doi={10.1109/ICASSP43922.2022.9746574}}

@inproceedings{huang2022e2e,
  title     = {{E2E} Segmenter: Joint Segmenting and Decoding for 
               Long-Form {ASR}},
  author    = {Huang, W. Ronny and Chang, Shuo-Yiin and Rybach, David 
               and Prabhavalkar, Rohit and Sainath, Tara N. and Allauzen, 
               Cyril and Peyser, Cal and Lu, Zhiyun},
  booktitle = {Proc. Interspeech},
  pages     = {4995--4999},
  year      = {2022},
  doi       = {10.21437/Interspeech.2022-38}
}

@article{stivers2009universals,
  title     = {Universals and cultural variation in turn-taking in conversation},
  author    = {Stivers, Tanya and Enfield, N. J. and Brown, Penelope and 
               Englert, Christina and Hayashi, Makoto and Heinemann, Trine and 
               Hoymann, Gertie and Rossano, Federico and {de Ruiter}, {Jan Peter} 
               and Yoon, Kyung-Eun and Levinson, Stephen C.},
  journal   = {Proceedings of the National Academy of Sciences},
  volume    = {106},
  number    = {26},
  pages     = {10587--10592},
  year      = {2009},
  publisher = {National Academy of Sciences},
  doi       = {10.1073/pnas.0903616106}
}

@inproceedings{narayanan2020recognizing,
  title={Recognizing Long-Form Speech Using Streaming End-to-End Models},
  author={Arun Narayanan and Rohit Prabhavalkar and Chung-Cheng Chiu and David Rybach and Tara N. Sainath and Trevor Strohman},
  journal={2019 IEEE Automatic Speech Recognition and Understanding Workshop (ASRU)},
  year={2019},
  pages={920-927},
  doi={DOI:10.1109/ASRU46091.2019.9003913}
}

@misc{hamming2025latency,
  title        = {{Voice AI Latency: What's Fast, What's Slow, and How to Fix It}},
  author       = {{Hamming AI}},
  year         = {2026},
  howpublished = {\url{https://hamming.ai/resources/voice-ai-latency-whats-fast-whats-slow-how-to-fix-it}},
  note         = {Accessed: 2026}
}

@article{cast2023,
title = {CAST: Context-association architecture with simulated long-utterance training for mandarin speech recognition},
journal = {Speech Communication},
volume = {155},
pages = {102985},
year = {2023},
issn = {0167-6393},
doi = {https://doi.org/10.1016/j.specom.2023.102985},
url = {https://www.sciencedirect.com/science/article/pii/S016763932300119X},
author = {Yue Ming and Boyang Lyu and Zerui Li}

}

@inproceedings{masumura2022contextual,
  title     = {Contextual-Utterance Training for Automatic Speech Recognition},
  author    = {G{\'o}mez Alan{\'\i}s, Alejandro and Drude, Lukas and Schwarz, Andreas 
               and Swaminathan, Rupak Vignesh and Wiesler, Simon},
  booktitle = {Proceedings of the IberSPEECH Conference},
  pages     = {26--30},
  year      = {2022},
  eprint    = {2210.16238},
  archivePrefix = {arXiv}
}

@inproceedings{machavcek2023whisper,
    title = "Turning Whisper into Real-Time Transcription System",
    author = "Mach{\'a}{\v{c}}ek, Dominik  and
      Dabre, Raj  and
      Bojar, Ond{\v{r}}ej",
    editor = "Saha, Sriparna  and
      Sujaini, Herry",
    booktitle = "Proceedings of the 13th International Joint Conference on Natural Language Processing and the 3rd Conference of the Asia-Pacific Chapter of the Association for Computational Linguistics: System Demonstrations",
    month = nov,
    year = "2023",
    address = "Bali, Indonesia",
    publisher = "Association for Computational Linguistics",
    url = "https://aclanthology.org/2023.ijcnlp-demo.3/",
    doi = "10.18653/v1/2023.ijcnlp-demo.3",
    pages = "17--24"
}

@misc{nemotron2024streaming,
  author       = {NVIDIA},
  title        = {{Nemotron-Speech Streaming EN 0.6B}},
  year         = {2024},
  publisher    = {Hugging Face},
  howpublished = {\url{https://huggingface.co/nvidia/nemotron-speech-streaming-en-0.6b}},
  note         = {Accessed: 2026-05-20}
}

@inproceedings{Xiao2024EfficientSL,
  title={Efficient Streaming Language Models with Attention Sinks},
  author={Guangxuan Xiao and Yuandong Tian and Beidi Chen and Song Han and Mike Lewis},
  booktitle={Proceedings ICLR},
  year={2024},
  url={https://arxiv.org/pdf/2309.17453}
}

@misc{silero2021vad,
  author       = {{Silero Team}},
  title        = {Silero {VAD}: Pre-trained Enterprise-Grade Voice Activity
                  Detector, Number Detector and Language Classifier},
  year         = {2021},
  publisher    = {GitHub},
  journal      = {GitHub repository},
  howpublished = {\url{https://github.com/snakers4/silero-vad}}
}

@article{romana2023,
title = {Automatic Disfluency Detection from Untranscribed Speech},
author = {Amrit Romana, Kazuhito Koishida, Emily Mower Provost},
year = {2023}
}

@inproceedings{gillick1989some,
  title     = {Some statistical issues in the comparison of speech recognition algorithms},
  author    = {Gillick, Laurence and Cox, Stephen J.},
  booktitle = {International Conference on Acoustics, Speech, and Signal Processing (ICASSP)},
  pages     = {532--535},
  volume    = {1},
  year      = {1989},
  address   = {Glasgow, Scotland},
  doi       = {10.1109/ICASSP.1989.266481},
  publisher = {IEEE}
}

\appendix

\begin{table*}[t]

\centering
\small
\renewcommand{\arraystretch}{1.1}
\begin{tabular}{ll l cc cc}
\toprule
& & & \multicolumn{2}{c}{\texttt{fastconformer-114m}} & \multicolumn{2}{c}{\texttt{nemotron-streaming-0.6b}} \\
\cmidrule(lr){4-5} \cmidrule(lr){6-7}
& & & Shape & Memory & Shape & Memory \\
\midrule
\multirow{3}{*}{Encoder State} & \multicolumn{2}{l}{Attention KV Cache} & $(17, 1, 70, 512)$ & 2.32\,MB & $(24, 1, 70, 1024)$ & 6.56\,MB \\
& \multicolumn{2}{l}{Convolution Cache} & $(17, 1, 512, 8)$ & 0.27\,MB & $(24, 1, 1024, 8)$ & 0.75\,MB \\
& \multicolumn{2}{l}{Valid Cache Length} & $(1,)$ & 8\,B & $(1,)$ & 8\,B \\
\midrule
\multirow{3}{*}{Decoder State} & \multicolumn{2}{l}{LSTM Recurrent State} & $(3, 640)$ & 7.5\,KB & $(10, 640)$ & 25\,KB \\
& \multicolumn{2}{l}{Last Token} & scalar & 8\,B & scalar & 8\,B \\
& \multicolumn{2}{l}{Decoded Length} & scalar & 8\,B & scalar & 8\,B \\
\midrule
Total & & & &2.59 MB & & 7.32 MB\\
\bottomrule
\end{tabular}
\caption{Model state components with the corresponding tensor shapes and memory footprint.}
\label{tab:state_components}
\end{table*}
\begin{table*}[t]
\centering
\small
\renewcommand{\arraystretch}{1.1}
\begin{tabular}{ll ccc}
\toprule
\textbf{Dataset} & \textbf{Model} & \textbf{$\Delta$\% (FS $\to$ RB)} 
& \textbf{MAPSSWE} & \textbf{Wilcoxon} \\
\midrule
\multirow{2}{*}{Switchboard} 
& \texttt{nemotron-streaming-0.6b} & 2.47 & $^{**}$ & $^{**}$ \\
& \texttt{fastconformer-114m} & 14.36 & $^{**}$ & $^{**}$ \\
\midrule
\multirow{2}{*}{CallHome} 
& \texttt{nemotron-streaming-0.6b} & 2.17 & $^{*}$ & $^{**}$ \\
& \texttt{fastconformer-114m} & 9.91 & $^{**}$ & $^{**}$ \\
\bottomrule
\end{tabular}
\caption{Relative WER reduction (\%) from FS to RB; statistical significance is assessed using the MAPSSWE and per-utterance Wilcoxon signed-rank tests. $^{*}$ indicates $p<0.05$ and $^{**}$ indicates $p<0.001$.}
\label{tab:sig_test}
\end{table*}
\begin{table*}[t]
\centering
\small
\begin{tabular}{ll cccc}
\toprule
Model & Dataset & SC&  RB ($k=1$) & RB ($k=5$) & RB ($k=10$) \\
\midrule
\multirow{4}{*}{\texttt{fastconformer-114m}} & CallHome (val)  &13.94& 13.82 & \textbf{13.70} & 13.74 \\
 & Switchboard (val) &9.95& 9.90 & \textbf{9.87} & 9.91 \\
 & CallHome (test) &14.02& 13.92 & \textbf{13.83} & 13.90 \\
 & Switchboard (test)  &10.26& 10.21 & \textbf{10.20} & 10.25 \\
\midrule
\multirow{2}{*}{\texttt{nemotron-streaming-0.6b}} & CallHome (val)  &13.83& 13.75 & 13.66 & \textbf{13.57} \\
 & Switchboard (val) &10.37& 10.30 & 10.05 & \textbf{10.02}\\
 & CallHome (test) &13.86& 13.71 & 13.62 & \textbf{13.52} \\
 & Switchboard (test)  &10.85& 10.80 & 10.68 & \textbf{10.62 }\\
\bottomrule
\end{tabular}
\caption{Comparison of overall WER (\%) across different token thresholds for RB.}
 \label{tab:RB}
\end{table*}

\begin{table*}[t]
\centering
\small
\renewcommand{\arraystretch}{1.1}
\begin{tabular}{ll cccc cccc}
\toprule
\multirow{2}{*}{\textbf{Model}} & \multirow{2}{*}{\textbf{Latency}} & \multicolumn{4}{c}{\textbf{CallHome}} & \multicolumn{4}{c}{\textbf{Switchboard}} \\
\cmidrule(lr){3-6} \cmidrule(lr){7-10}
& & \textsc{fs} & \textsc{cs} & \textsc{sc} & \textsc{rb} & \textsc{fs} & \textsc{cs} & \textsc{sc} & \textsc{rb} \\
\midrule
\multirow{4}{*}{\texttt{fastconformer-114m}} & 80 ms & 15.52 & 15.64 & 14.47 & \textbf{14.19} & 12.21 & 12.81 & 10.57 & \textbf{10.49} \\
 & 160 ms & 15.43 & 15.18 & 14.02 & \textbf{13.83} & 11.97 & 12.36 & 10.26 & \textbf{10.20} \\
 & 240 ms & 15.45 & 15.37 & 14.06 & \textbf{13.98} & 12.04 & 12.66 & 10.28 & \textbf{10.27} \\
\cmidrule(lr){3-10}
 & 560 ms & 14.75 & 15.04 & 13.49 & \textbf{13.44} & 11.44 & 12.06 & \textbf{10.04} & \textbf{10.04} \\
\midrule
\midrule
\multirow{4}{*}{\texttt{nemotron-streaming-0.6b}} & 80 ms & 14.51 & 14.86 & 14.35 & \textbf{14.16} & 11.44 & 11.88 & 11.5 & \textbf{11.17} \\
 & 160 ms & 13.82 & 14.29 & 13.86 & \textbf{13.52} & 10.90 & 11.40 & 10.85 & \textbf{10.62} \\
 & 240 ms & 13.97 & 14.67 & 13.83 & \textbf{13.47} & 10.97 & 11.61 & 11.01 & \textbf{10.73} \\
\cmidrule(lr){3-10}
 & 560 ms & \textbf{12.82} & 14.15 & 13.77 & 13.33 & \textbf{10.08} & 11.19 & 10.86 & 10.67 \\
 \midrule
\bottomrule
\end{tabular}
\caption{Comparison of overall WER (\%) across different state-management strategies and latency settings.}

\label{tab:lat}
\end{table*}
\section{Appendix}
\label{sec:app}
\subsection{Inference Setup}
\label{app:inf}
All experiments were conducted using Python 3.10.19 and the NeMo ASR toolkit version 2.7.0, with all model checkpoints obtained from Hugging Face. Inference was performed on NVIDIA L40 GPUs. To emulate real-world streaming deployment conditions, all evaluations were carried out independently with a batch size of 1. Both streaming ASR models employ deterministic greedy decoding during inference; consequently, all reported results are obtained from a single run.

\subsection{Architecture Details}
\label{app:arch}
Both models operate on 16\,kHz audio and produce subword token sequences. The smaller \texttt{fastconformer-114m} model consists of a 17-layer FastConformer encoder with a hidden dimension of 512, resulting in approximately \textsc{114M} parameters and a model size of 435\,MB. The model is trained using a hybrid RNNT-CTC objective on the datasets described in \citet{noroozi2024stateful}; throughout our experiments, we use the RNNT decoder for inference.
The \texttt{nemotron-streaming-0.6b} model employs a larger 24-layer FastConformer encoder with a hidden dimension of 1024, yielding approximately 0.6B parameters across the encoder and decoder and a memory footprint of approximately 2.25\,GB. Unlike \texttt{fastconformer-114m}, it is trained exclusively with an RNNT objective on the data described in \citet{nemotron2024streaming}.

{\paragraph{Model state.} For both the streaming models, the model state consists of a tuple of six components, which can be further divided into two groups: encoder (FastConformer) state components and decoder (RNNT) state components. The encoder state comprises two continuous cache tensors: (i) the attention KV cache (shape: \texttt{[n\_layers, batch\_size, left\_context\_len, hidden\_dim]}) and (ii) the convolution cache (shape: \texttt{[n\_layers, batch\_size, hidden\_dim, conv\_subsample]}) for cache-aware streaming, plus (iii) a small counter tracking valid cache length. The decoder state holds (iv) the RNNT predictor's recurrent LSTM states, (v) its last output token, and (vi) the decoded length for the token sequence emitted so far. Table \ref{tab:state_components} provides the shapes and memory footprint for each of the state components for both the streaming models.}

\subsection{Dataset Statistics}
\label{app:data}
 Switchboard \cite{godfrey1992swb,Deshmukh1998ResegmentationOS}
and CallHome \cite{canavan1997callhome} are two of the most commonly used telephonic dialogue benchmarks. Switchboard consists of telephone conversations between strangers discussing assigned topics, resulting in relatively structured dialogues with clearer turn-taking behavior. In contrast, CallHome contains conversations between friends and family members, making the speech more spontaneous, informal, and conversationally dense, with more frequent backchannels, interruptions, and topic shifts. Consequently, CallHome is generally considered the more challenging benchmark for ASR systems and provides a useful testbed for evaluating robustness to natural conversational phenomena.

To approximate the end-of-turn detection behavior employed in commercial voice agents \cite{hamming2025latency}, we combine consecutive utterances from the same speaker separated by less than 1\,s of silence and treat them as a single conversational turn. The CallHome test set comprises 40 channels containing 1,335 turns, whereas the Switchboard test set \cite{romana2023} contains 214 channels and 5,822 turns; each channel corresponds to approximately 5 minutes of conversational audio. The average turn length is 14.2 words for the CallHome corpus and 16.5 words for the Switchboard corpus.
\begin{table*}[t]
\centering
\small
\begin{tabular}{llcccc}

\toprule
\textbf{Model} & \textbf{Category} & \textbf{FS} & \textbf{CS} & \textbf{SC} & \textbf{RB} \\
\midrule
\multirow{4}{*}{\texttt{fastconformer-114m}} 
& Overall    & 14.71 & 15.98 & 13.84 & \textbf{13.59} \\
& First Word & 49.15 & 45.92 & 35.05 & \textbf{32.50} \\
& After Long Pause   & 18.17 & 23.33 & 17.13 & \textbf{16.97} \\
& After Short Turn   & 16.61 & 19.41 & 15.63 & \textbf{15.39} \\
\midrule
\multirow{4}{*}{\texttt{nemotron-streaming-0.6b}} 
& Overall    & 14.79 & 15.74 & 14.66 & \textbf{14.28} \\
& First Word & 33.15 & 37.72 & 32.95 & \textbf{30.72} \\
& After Long Pause  & 17.71 & 19.68 & 17.75 & \textbf{17.19} \\
& After Short Turn   & 16.58 & 18.71 & 16.92 & \textbf{16.01} \\
\bottomrule
\end{tabular}
\caption{Performance comparison of different state-management strategies on Switchboard with 160\, ms latency when turn boundaries are derived from Silero-VAD. Metric definitions are identical to those used in Table \ref{tab:phenomena_analysis_v3}.}
\vspace*{-1em}
\label{tab:vad_timings}
\end{table*}

\subsection{Statistical Significance Testing}
\label{app:sigtest}
{We assess the statistical significance of the difference between FS and RB in \textit{Overall WER} using the Matched-Pair Sentence-Segment Word Error (MAPSSWE) test, with the Wilcoxon signed-rank test on utterance-level WER differences serving as an additional confirmation \cite{gillick1989some}.
MAPSSWE pools error counts across all segments, with utterances weighted by their number of reference words. In contrast, the Wilcoxon test ranks signed per-utterance WER differences, assigning equal weight to each turn.
As shown in Table \ref{tab:sig_test}, improvements from our methods are statistically significant under both tests, with $p<10^{-3}$ for all but one result. The results for \texttt{nemotron-streaming-0.6b} with the CallHome dataset are statistically significant with $p<0.05$ for the MAPSSWE test and $p<10^{-3}$ for the per-utterance Wilcoxon test. Notably, the Wilcoxon test, which considers individual utterance-level WER improvements, consistently yields higher statistical significance than the MAPSSWE test.

Finally, our primary contribution is improving recognition at turn onsets rather than overall utterance-level WER. We therefore view these significance tests as secondary evidence that our state-management strategies improve, rather than degrade, overall transcription quality, while the turn-onset analyses presented in Section \ref{sec:res} provide the primary evidence for our contribution.}

\subsection{Rollback Threshold}
\label{app:eval}
The State Rollback (RB) strategy described in Section \ref{subsec:state} uses an empirically tuned threshold $k$. Specifically, if the preceding turn contains less than $k$ tokens, the model rolls back to the state saved after the most recent substantial utterance, defined as one for which the ASR model emitted at least $k$ tokens. This allows RB to discard context from short turns, such as backchannels, while preserving context from longer utterances. In this study, we evaluate coarse thresholds $k \in \{1,5,10\}$ and select the configuration with the best overall validation WER for the results presented in Section \ref{sec:res}. As shown in Table \ref{tab:RB}, this corresponds to $k=5$ for \texttt{fastconformer-114m} and $k=10$ for \texttt{nemotron-streaming-0.6b}. {The same optimal thresholds are observed across both datasets and their respective validation and test splits, suggesting that an appropriate value of $k$ can be selected through validation-set tuning for real-world deployment.} These results are consistent with our earlier observation that \texttt{nemotron-streaming-0.6b} is more sensitive to performance degradation following short turns and backchannels under SC, motivating a higher rollback threshold. Finally, we observe that RB consistently outperforms SC across all settings for all the evaluated thresholds.
\subsection{Latency Analysis}
\label{app:latency}
Both streaming ASR models are trained with a left-context window of $L=70$ frames, corresponding to 5.6\,s of acoustic memory, and lookahead values of $R\in\{0,1,6,13\}$, corresponding to latencies of 80\,ms, 160\,ms, 560\,ms, and 1120\,ms, respectively \cite{nemotron2024streaming}.  Although inference is possible with arbitrary positive values of $\{L,R\}$, we observe the best performance when using the configurations seen during training. All the results mentioned in Section \ref{sec:res} use $L=70,R=1$, which is the best-performing model training configuration for conversational voice agent applications that need to transcribe with a latency budget below 250\,ms \cite{hamming2025latency}.
To evaluate the robustness of our state-management strategies to latency constraints, we test them at target latencies of 80\,ms and 240\,ms. Table \ref{tab:lat} reports the performance of all strategies across latency settings for both models with ground-truth timings. We additionally include results at 560\,ms, a latency setting used during model training. While unsuitable for real-time voice agents, this setting serves as a useful reference for understanding the impact of increased lookahead on performance.

The results in Table \ref{tab:lat} show that the trends observed at 160\,ms latency remain consistent across other latency settings, with RB consistently outperforming the alternative strategies across all models, datasets, and voice-agent-compatible latencies. Comparison with the reference reveals that reducing the latency from 560 ms to 160 ms results in less than a 6\% increase in overall WER for both models under RB. For \texttt{fastconformer-114m}, overall WER for RB with a latency budget of 160\,ms is actually lower than that of FS with 560\, ms latency for both datasets. The larger context window available at 560\,ms latency reduces the need for prior-turn contextual information, allowing FS to outperform the other state-management strategies for \texttt{nemotron-streaming-0.6b}.

\subsection{VAD-derived Turn Boundaries}
\label{app:vad}
 To mimic realistic deployment settings, we use Silero-VAD to detect the turn onsets and ends.  Consistent with the ground-truth timing setup, the VAD uses a minimum pause threshold of 1\,s and extends the turn ends by 500\,ms to accommodate delayed token emissions. Any tokens produced outside the resulting speech regions are excluded from evaluation. Most Silero-VAD parameters were kept at their default settings \cite{silero2021vad}, with some modification: the minimum turn length was reduced to 100\,ms to better capture short turns. 

To reduce crosstalk in the VAD-derived speech regions, we compare speech segments across opposite channels and remove long overlapping regions (>500\,ms) from the channel with lower signal energy. This heuristic effectively filters out crosstalk while preserving short backchannels and genuine overlaps.
With these settings, VAD detects speech activity covering approximately 98.8\% of the words in Switchboard. We further observe that the VAD-derived boundaries are slightly more permissive than the ground-truth annotations, increasing the average detected speech duration from 139\,s to 147\,s per channel, potentially exposing the ASR model to residual crosstalk and telephony-channel noise despite our filtering procedure.

{Under this setup, each VAD-detected speech segment is treated as a separate turn. Each reference word is attributed to the turn containing its temporal midpoint, with words falling outside all detected segments assigned to the nearest turn. For WER calculation, we consider only turns containing at least one attributed reference word. This analysis is limited to Switchboard, as CallHome does not provide the word-level timing annotations required for turn attribution \cite{canavan1997callhome}.}

The results for Switchboard with VAD timings for a setup with 160\,ms latency are provided in Table \ref{tab:vad_timings}. The overall WER with VAD-derived timings is consistently higher than with ground-truth timings. This degradation is likely
caused by a combination of residual crosstalk leakage as well as missed words and speech segments resulting from imperfect VAD detection. The overall trends are quite similar to the  experiments using ground-truth timings, with our strategies outperforming the existing methods for both the models.

\subsection{Use of Existing Artifacts}
This work exclusively uses publicly available datasets and open-source models. The artifacts were employed in a manner consistent with their intended use. To support reproducibility, we have released 
the software and implementation details associated with this work.

\subsection{AI Assistant Usage}
The authors used generative AI tools to assist with grammar correction and improving the readability of the manuscript. The AI tools were not used to generate research ideas, conduct experiments, analyze results, or draw scientific conclusions. All technical content and final manuscript revisions were reviewed and approved by the authors.

\end{document}